\documentclass[letterpaper, 10 pt, conference]{ieeeconf}  

\IEEEoverridecommandlockouts                              

\usepackage{graphicx}
\usepackage[table]{xcolor}
\usepackage{subcaption}

\title{\LARGE \bf
Scale Optimization for Full-Image-CNN Vehicle Detection
}

\author{Yang Gao, Shouyan Guo, Kaimin Huang, Jiaxin Chen, Qian Gong, Yang Zou, Tong Bai and Gary Overett$^{a}$
\thanks{$^{a}$Authors are with the Joint Institute of Engineering, Sun Yat-sen University -- Carnegie Mellon University,
  510006 Guangdong, China
        {\tt\tiny yanggao;shouyang;kaiminh1;jiaxinc;qgong1;yzou2;tongb;overett@cmu.edu}}%
}

\begin{document}

\maketitle
\thispagestyle{empty}
\pagestyle{empty}

\begin{abstract}

  Many state-of-the-art general object detection methods make use of shared full-image convolutional features (as in Faster R-CNN). This achieves a reasonable test-phase computation time while enjoys the discriminative power provided by large Convolutional Neural Network (CNN) models. Such designs excel on benchmarks\footnote{MS COCO \cite{lin2014microsoft}, VOC \cite{everingham2010pascal}, ImageNet \cite{russakovsky2015imagenet}} which contain natural images but which have very unnatural distributions, i.e. they have an unnaturally high-frequency of the target classes and a bias towards a ``friendly'' or "dominant" object scale. In this paper we present further study of the use and adaptation of the Faster R-CNN object detection method for datasets presenting natural scale distribution and unbiased real-world object frequency. In particular, we show that better alignment of the detector scale sensitivity to the extant distribution improves vehicle detection performance. We do this by modifying both the selection of Region Proposals, and through using more scale-appropriate full-image convolution features within the CNN model. By selecting better scales in the region proposal input and by combining feature maps through careful design of the convolutional neural network, we improve performance on smaller objects. We significantly increase detection AP for the KITTI dataset car class from 76.3\% on our baseline Faster R-CNN detector to 83.6\% in our improved detector. 

\end{abstract}

\section{INTRODUCTION}

Recently, several design variations for object detection using region based convolutional neural networks have generated
state-of-the-art performance against traditional many-class object detection benchmarks \cite{girshick2014rich, girshick2015fast, fasterrcnn, redmon2016you}. These datasets typically present target objects with unnaturally high target object
frequency and ``friendly'' or dominant scale. This is a natural consequence of the data collection methodology casting a
prior bias by seeking images specifically containing a chosen set of target classes \cite{everingham2010pascal, lin2014microsoft}. i.e. the benchmark images were chosen from a larger pool of available images because they contain
  examples of one (often more) instances of a chosen class and furthermore contain these examples usually at
  significant (often dominant) scale so as to be easily labeled.

A consequence of such popular benchmarking is that leading object detectors play to this benchmark through design
choices. In particular, designing detectors for such datasets requires only moderate attention to both the detectors
scale-invariance and the  \emph{much lower frequency} of objects in the real-world. Consider, the case of scanning a Pascal-VOC trained Faster-RCNN detector over a
random selection of Flickr images (https://www.flickr.com/), detector precision for the person class might be acceptable while the precision 
for a lower frequency class such as horses might be very poor, with the output being dominated by false positives.

\begin{figure}[thpb]
  \centering

  \begin{subfigure}[t]{.5\linewidth}
    \centering
    \includegraphics[width=.95\linewidth]{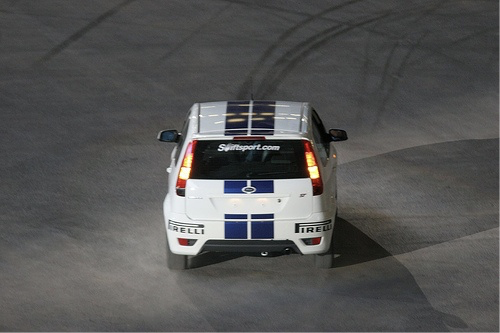}
    \caption{VOC2007 Image}
    \label{fig:sub1}
  \end{subfigure}%
  \begin{subfigure}[t]{.5\linewidth}
    \centering
    \includegraphics[width=.95\linewidth]{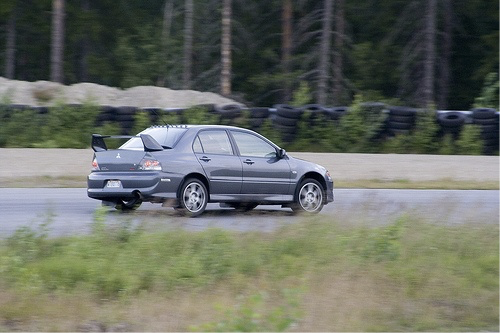}
    \caption{VOC2007 Image}
    \label{fig:sub2}
  \end{subfigure}

  \begin{subfigure}[t]{\linewidth}
    \centering
    \includegraphics[width=0.98\linewidth]{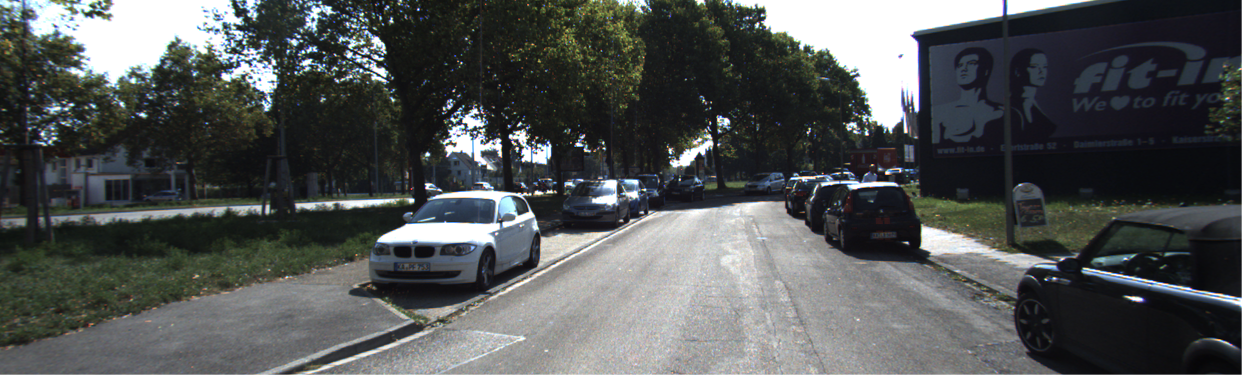}
    \caption{KITTI Dataset Image}
    \label{fig:sub3}
  \end{subfigure}

  \begin{subfigure}[t]{.5\linewidth}
    \centering
    \includegraphics[width=\linewidth]{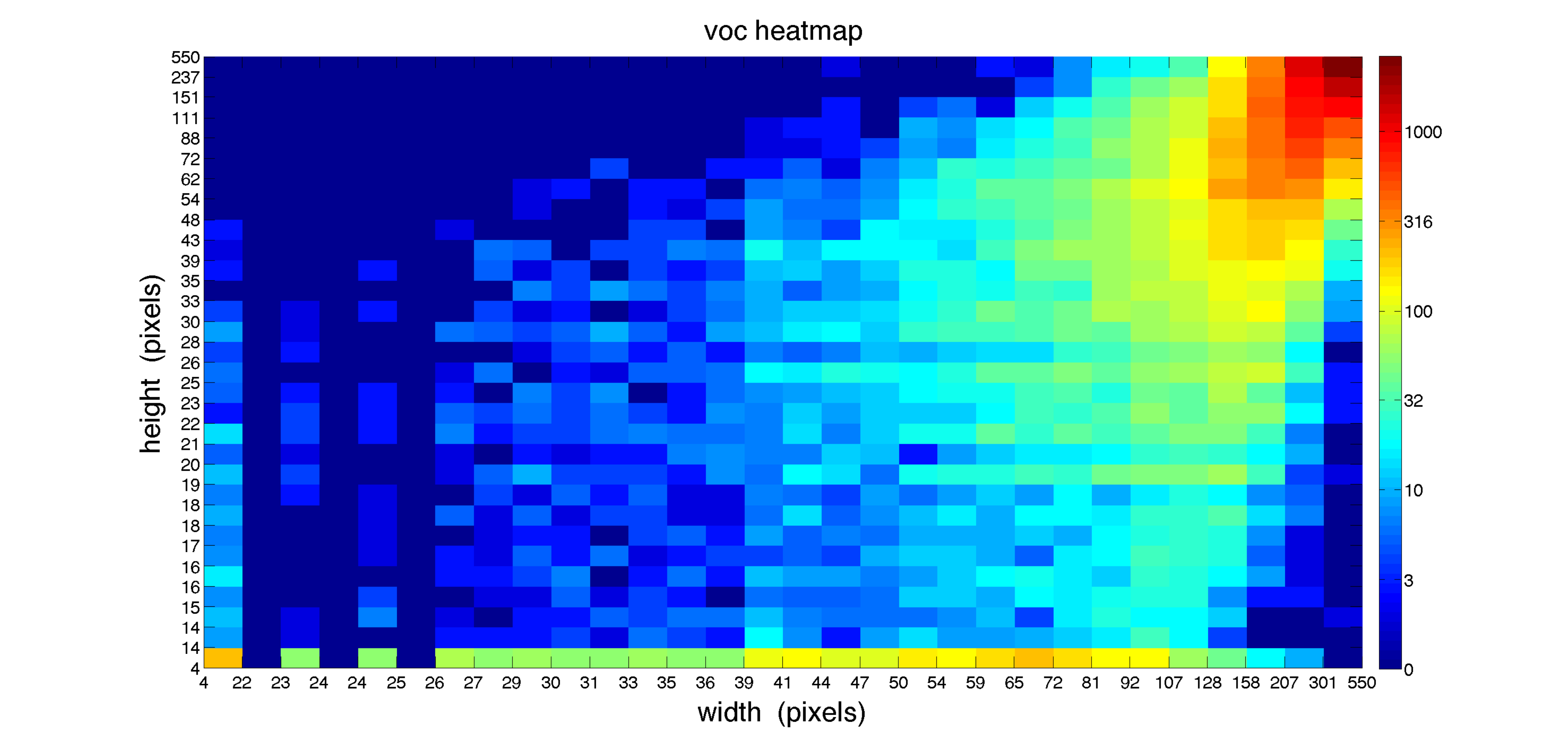}
    \caption{VOC2007 Scale Distribution}
    \label{fig:sub4}
  \end{subfigure}%
  \begin{subfigure}[t]{.5\linewidth}
    \centering
    \includegraphics[width=\linewidth]{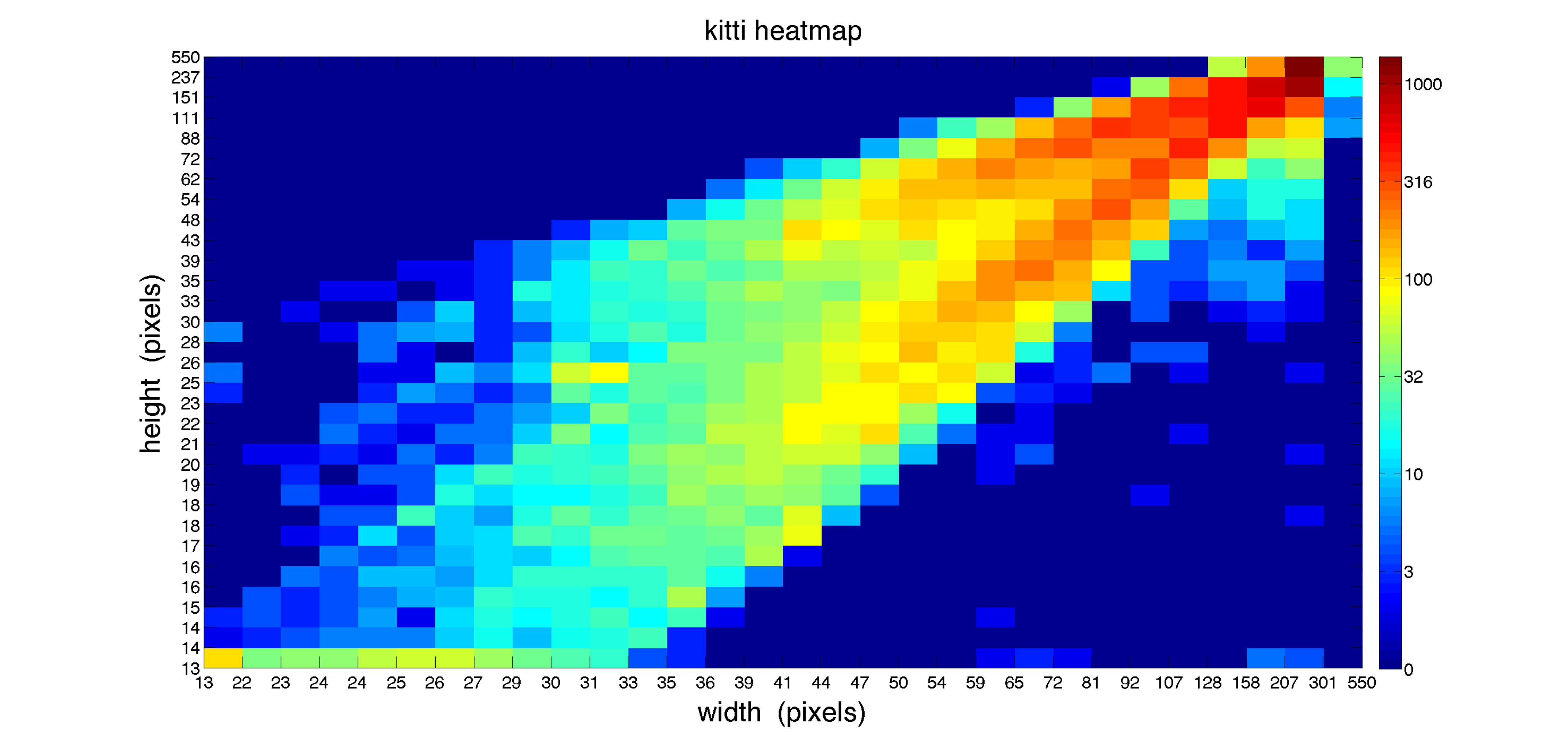}
    \caption{KITTI Scale Distribution}
    \label{fig:sub5}
  \end{subfigure}

  \caption{VOC vs KITTI Benchmark Object Scale Comparison. Here we see the KITTI Benchmark presents a more
    ``domain-natural'' distribution of object scales. As the vehicle moves through the scene the car scales following the
    expected distribution of an object during forward motion. We also observe 2 distinct 'aspect-modes' caused by the presence of front/rear and side vehicles perspectives. The VOC dataset is dominated by larger objects that consume a large portion of the image scene.}
  \label{kittivocscales}
\end{figure}

On the other hand, domain specific object detection benchmarks such as the KITTI Vision Benchmark Suite \cite{geiger2013vision} present what we might call a more ``domain-natural'' distribution. That is,
for the vehicle detection domain (i.e. the forward facing road scene from a driving vehicle), the scale distribution
found in the benchmark represents a more typical presentation of scales. The scale of vehicles in the scene presents
naturally according to the (usually) forward motion of the vehicle. A small/distant vehicle is as likely to present in
the benchmark as a larger/closer vehicle. Figure \ref{fig:sub4} and Figure \ref{fig:sub5} compare the distribution of car images in the VOC and KITTI
datasets respectively.

While leading object detection methods were designed somewhat specifically for more unnatural general object detection
challenges, we believe that many of the underlying design choices used in these methods are valuable. In particular, the
\emph{shared full-image convolution features} approach of the Faster R-CNN method \cite{fasterrcnn} shows a way forward
in object detection where it is possible to use the more powerful large CNN based models while not suffering all of the
very large computational burden \cite{girshick2014rich} inherent in moving from prior detector designs such as attentional cascades
of fast hand-crafted features \cite{viola2001rapid}. The key insight here is that it is possible to apply the heavy CNN
computation once over the entire image to produce a feature description of the content which can be extracted and
analysed locally for objects within the image without evaluating the CNN on many thousands of local image
patches. However, as we demonstrate (See Section \ref{optim}) the Faster R-CNN design - though brilliant - has a cost in terms of
the scale invariance of the image description obtained.

In this work, we propose maintaining these key design choices given by the Faster R-CNN method and explore the further
optimization of the approach for a domain-natural object detection distribution, specifically the KITTI Vehicle
Detection Benchmark. However, rather than following the standard benchmarking scheme which is somewhat focused on
improving the mean average precision (MAP) against leading benchmark contenders, we specifically study the response of
method variations to images at different scales. Therefore, whereas the KITTI dataset has been divided into 3 subsets
(easy, medium, \& hard) based on scale, occlusions, and truncation, we specifically study the effect of scale and the inclusion of shallower layer feature maps in our
method on the overall representation of KITTI dataset.
\par
Through our study, we found that the careful selection of smaller anchor boxes and shallower features can greatly improve the detection accuracy of vehicles in the KITTI dataset.

\section{RELATED WORK}

The issue of image scale has long been important in object detection. An early and obvious approach is the use of a
scale-space pyramid and windowing \cite{szeliski2010computer} to allow a detector to only consider the problem at a single window scale. This can be particularly, effective
when the detector is an attentional cascade \cite{viola2001rapid} or a relatively fast support vector machine approach with
suitably fast features \cite{dalal2005histograms}. However, such an approach can be challenging to integrate with today's
powerful neural network models which do not yield such computationally minimalistic features. For this reason, some systems have chosen
to use modern neural network designs only in the `tail-end' of their detection cascades where the average per-image
computation burden is low \cite{li2015convolutional, verma2015pedestrian}

Yet, to use neural networks only at the tail-end of a detection cascade is to miss out on some of the benefits they
offer. It has been shown in recent times that hard crafted features simply tend to miss out on some of the general
discriminative power available to CNN's through a pool of multi-layer co-optimized (usually through gradient descent)
feature sets. Furthermore, CNN features excel in moving up the discriminative value chain from general low-level
features for image understanding (edges, textures) to through to higher-level features-of-features (eyes, wheels,
vehicle-grills, etc) and very importantly, when compared to cascade approaches, \emph{they share} these lower-level features between object classes.

The primary ancestor of modern CNN based object detectors, R-CNN \cite{girshick2014rich} combines a leading
classification method, AlexNet \cite{krizhevsky2012imagenet}, with a sparse region proposal \cite{uijlings2013selective, zitnick2014edge} method
which provided a set of candidate image sub-regions for classification according to target object classes. This produced
state-of-the-art performance but with very high computational cost as the number of region proposals was often
significant and each sub-region of the image required separate processing by the CNN.

The computational cost was significantly reduced in the Fast R-CNN \cite{girshick2015fast} approach. The main contribution of this work is to propose the idea of sharing the feature map of entire images for various region proposals. Specifically, Fast R-CNN first computes the feature maps for the whole image and extracts the region of features according to the "objectness" region proposal method. Since different proposals from an image can make use of the same feature map, we do not have to compute feature maps for \emph{every} proposal separately. The resulting region of interest feature map then takes the place of the last convolutional layer's feature output in the standard R-CNN to classify all the region proposals. This is followed by a bounding box regression to achieve accurate bounding box coordinates. This strategy greatly reduces repeat computation from overlapping regions. However, both R-CNN and Fast R-CNN use traditional region detection methods like Selective Search \cite{uijlings2013selective} to generate region proposals. This method is computationally expensive and becomes a bottleneck for fast, or real-time processing.

Evolved from R-CNN and Fast R-CNN, Ren proposed a Faster R-CNN \cite{fasterrcnn} approach consisting of
the Fast R-CNN method and a Region Proposals Network (RPN) sharing \emph{the same} CNN features. Faster R-CNN optimizes the region proposal process by introducing a Region Proposal Network (RPN), which improves the computational speed and proposal quality. It explores the capability of a sliding window and Fast R-CNN combined neural network for generating the objectness region proposals. The incorporation of the RPN method in the detection framework takes fuller advantage of the GPU, greatly improving the computational speed. Furthermore, the common structure of the RPN method can be utilized in the speeding up of training by sharing the parameters with the following proposal classification network - Fast R-CNN.

\par
The Faster R-CNN design extracts features from high-level convolution layers. For example, conv5 in ZF-Net \cite{zeiler2014visualizing}
which presents a downsampling factor of 32 \cite{geiger2013vision, fasterrcnn}. Consequently, the receptive field corresponding to
the original image is larger than smaller target vehicles within the KITTI dataset ($171^2$ pixels for ZF-Net). Intuitively,
the large receptive field introduces unrelated object and background information which dilutes the discriminative power
of the conv5 `feature description'. The result being that tiny objects often cannot be correctly represented or
detected.

 \begin{figure}[ht]
   \centering
   \includegraphics[width=.75\linewidth]{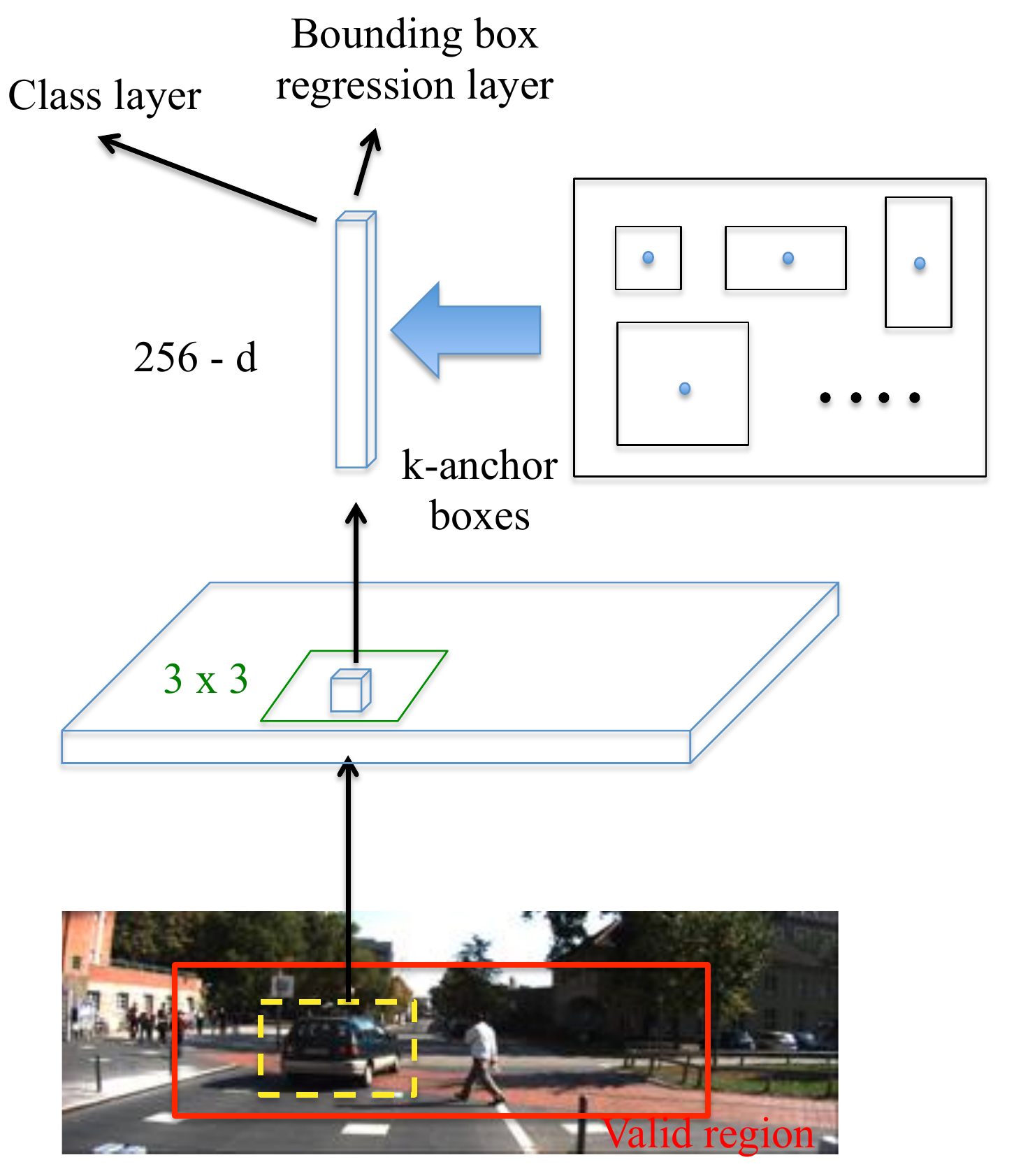}\\
   \caption{Faster R-CNN multi-task learning with different anchor box scales / ratios. A full-image convolution is used to process the input image into a feature map. A local region of this feature map is reduced to a 256-d vector which is used with anchor box region suggestions to infer both class/non-class detections and bounding box refinement relative to the input anchor box suggestions. We show that an appropriate matching of anchor box scales to the dataset distribution is important for the success of the method.}\label{fast-rcnn}
 \end{figure}

\par
It's easy to see that good region proposals are vital for efficient object detection. Fast R-CNN uses a traditional
Selective Search \cite{uijlings2013selective} to generate region proposals. This method is computationally expensive and is a
bottleneck for real-time processing. Faster R-CNN developed a Region Proposal Network to address this problem by
exploring the capability of a neural network for generating the proposals and sharing the parameter of RPN with Fast R-CNN to further improve the training speed. The improvement of the quality of proposed region is very crucial in the whole detection system.



Faster R-CNN's RPN layer utilizes a 3x3 sliding window approach over the convolution output in order to generate a set of
object proposals. This is done by summarizing the convolution output into a single fixed length (e.g. 256-dimensional)
description at each location through the application of a $3 \times 3$ convolution followed by a $1 \times 1$ 
convolution. This fixed length vector is then passed to a box-classification and box-regression layer. The task of the
box-classification layer is to determine the objectness of given region while the box-regression layer suggests offsets to the actual object location \emph{relative} to the region given as an ``anchor'' box.


Key to our analysis, this bounding box-classification and box-regression is computed \emph{relative} to a set of $k$
so-called anchor boxes presenting \emph{``suggested''} scales and aspect ratio variations at each location. The original
method used 3 scales and 3 aspect ratios in combination to yield $k=9$ anchors to the box-classification and
box-regression layers. Presumably, the original authors used some degree of empirical optimization against \emph{their}
chosen datasets in choosing these anchor box values (specifically box areas of [$128^2$,$256^2$,$512^2$] and aspect
ratios of [1:2,1:1,2:1]). As we will see, these anchor boxes are key to the success of the method across different domains. Ideally, anchor boxes should suggest the approximate location, scale, and aspect ratio of the objects we want to detect while ``suggesting" a minimal'' number of non-object regions. So the ratios and scales of these anchor boxes are very important.





Given these prior contributions, it is not surprising that others have already attempted to modify the Faster R-CNN
method for class-specific object detection. Notable, examples include pedestrian detection \cite{zhang2016faster} where hard background images and scale issues are identified as a key difficulty for
the existing approach, as well as for vehicle detection \cite{fan2016closer} where
the outer training and test parameters of the Faster R-CNN approach were explored. Given, a previous study for vehicle
detection using Faster R-CNN it is important for us to draw a distinction. The work of Q. Fan et al \cite{fan2016closer} did not consider issues relating to the scale of objects with reference to the 
algorithm internals, such as the anchor box method or the use of features pooled from different layers of the
CNN. Rather, they produced an informative exploration of issues of training and test input image size, the number of region
proposals used, and the training method.

In contrast, we will explore the internal algorithm of Faster R-CNN especially focusing on the scale of anchor boxes and features from different layers.

\section{OPTIMIZING FASTER R-CNN FOR DOMAIN-NATURAL VEHICLE SCALES}
\label{optim}

\subsection{Anchor Box Optimization}

To detect smaller objects, smaller anchor box proposals are needed to specifically address the larger presence of smaller objects. The original method used 3 anchor box scales of [$128^2$,$256^2$,$512^2$] pixels in area. Given the higher frequency of small vehicles (See Figure \ref{fig:sub5}) in the Kitti Vision Benchmark, we add two smaller scales in the anchor boxes generation process to cover the high-frequency interval of the dataset between 30-60 pixels in width. This yields boxes of [$32^2$,$64^2$,$128^2$,$256^2$,$512^2$] pixels in area. Since we have 3 ratios, the number of anchor boxes of each location is 15. The results show the usage of smaller anchor box can significantly increase the test AP (See Section \ref{anchorboxresults}).





\par

\subsection{CNN Optimization for Smaller Objects}

In addition to varying the number, scale, and distribution of anchor boxes we explore the possibility of changing the
actual network design of the full-image convolution layers. In particular, the very large receptive field of the
existing approach means that the convolutional feature map aggregates image information over a large area. For small objects, this leads to a dilution of the object information as background non-object information may dominate the feature. The large receptive field is
derived directly from the size of the convolution kernels used and the number of layers in the network. Lower
layers will have smaller receptive fields than layers above them. Therefore, we explore 3 networks redesigned based on
the original ZF-Net used in the original Faster R-CNN paper. In each case, we try to allow for the use of lower level more fine-grained scale information in the input image.

Higher-level features have a larger receptive field. As the features contain more global information, the smaller scale information is lost along with the objects precise position information. So while the
higher level feature map may be more descriptive, the high-level representation may not be suitable for accurate prediction of small objects. We have proposed three possible methods extending the final convolutional layer to get more fine-scale features.

\par
\begin{figure}[ht]
  \centering
  \includegraphics[width=\linewidth]{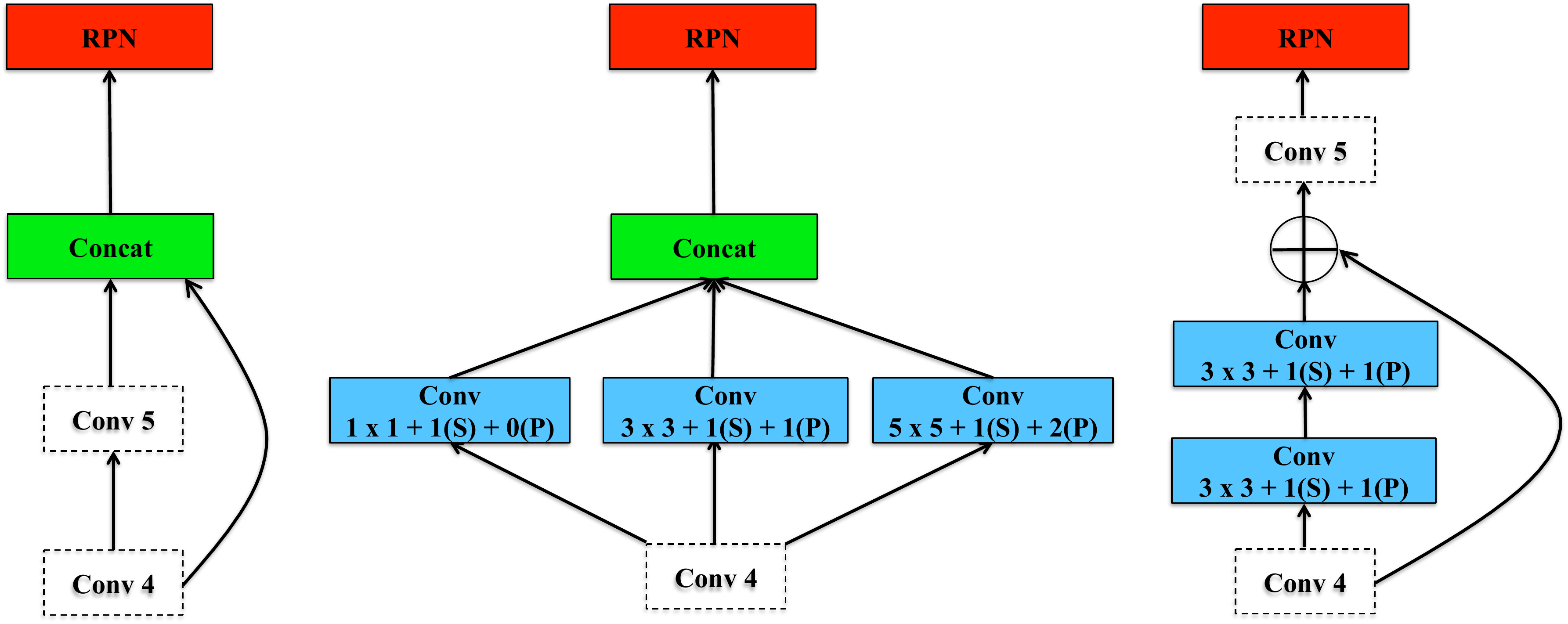}\\
  \caption{The proposed network architecture modifications. The dotted-line rectangle means the layers are initialized
    by a pre-trained model. The solid-line rectangle means the layers are trained from scratch. (a. left) The combinations of
    conv4 and conv5 feature maps in a multi-layer network. ZF$_{ml}$ (b. middle) The combinations of multi-scale convolution layers. ZF$_{ms}$ (c. right) The residual network
    embedding ZF$_{res}$.}
    \label{modification}
\end{figure}

\textbf{Multi-layer proposal. (ZF$_{ml}$)} 
Here we leverage the idea from DeepID1 \cite{sun2014deep}, concatenating the feature maps from conv4 and conv5, which have different receptive fields. i.e. conv4 is better placed to detect smaller objects than conv5. The resulting feature maps are fed to the RPN layer, as shown in Figure \ref{modification}. Since the combined features contain information gathered over a range of receptive scales, it yields better proposals and improves the detection performance faster during training. However, we find that when both networks are fully trained, the performance of Multi-layer proposal network is similar to the baseline model. 
\par
\textbf{Multi-scale proposal. (ZF$_{ms}$)} Inspired by the GoogleNet \cite{szegedy2015going}, we add multiple scale convolutional kernels (1$\times$1, 3$\times$3, 5$\times$5) to the conv4 layer and concatenate them to the conv5 layer. This has the advantages of multiscale convolutional fields each optimized for different scale information in the input image. The resulting feature maps are fed to the RPN layer, as shown in Figure \ref{modification}. The combined features contain different information of different convolutional scales yielding better object proposals and improved detection performance at the first stage of training. The final detection AP improves upon the baseline performance but not very significantly.

\par
\textbf{Residual block embedding. (ZF$_{res}$)} Deep Residual Networks \cite{he2016deep} have emerged as a state-of-the-art deep neural network architecture. Accordingly, we add a residual block between the conv4 and conv5 layers, as shown in Figure \ref{modification} (c). In the main branch, we apply two $3 \times 3$ convolution filters sequentially to the conv4 feature map while using an identity map in the shortcut branch. We add the two corresponding outputs together giving us a deeper layer of features while avoiding the loss of fine-scale information. The result is the input of the RPN layer. We expect the residual block embedding could boost the performance due to its more scale-flexible representation.



\begin{figure}[ht]
  \centering
  \includegraphics[width=\linewidth]{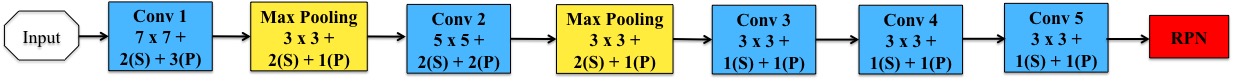}\\
  \caption{The baseline network structure.}\label{zfnet}
\end{figure}

\begin{figure}[ht]
  \centering
  \includegraphics[width=\linewidth]{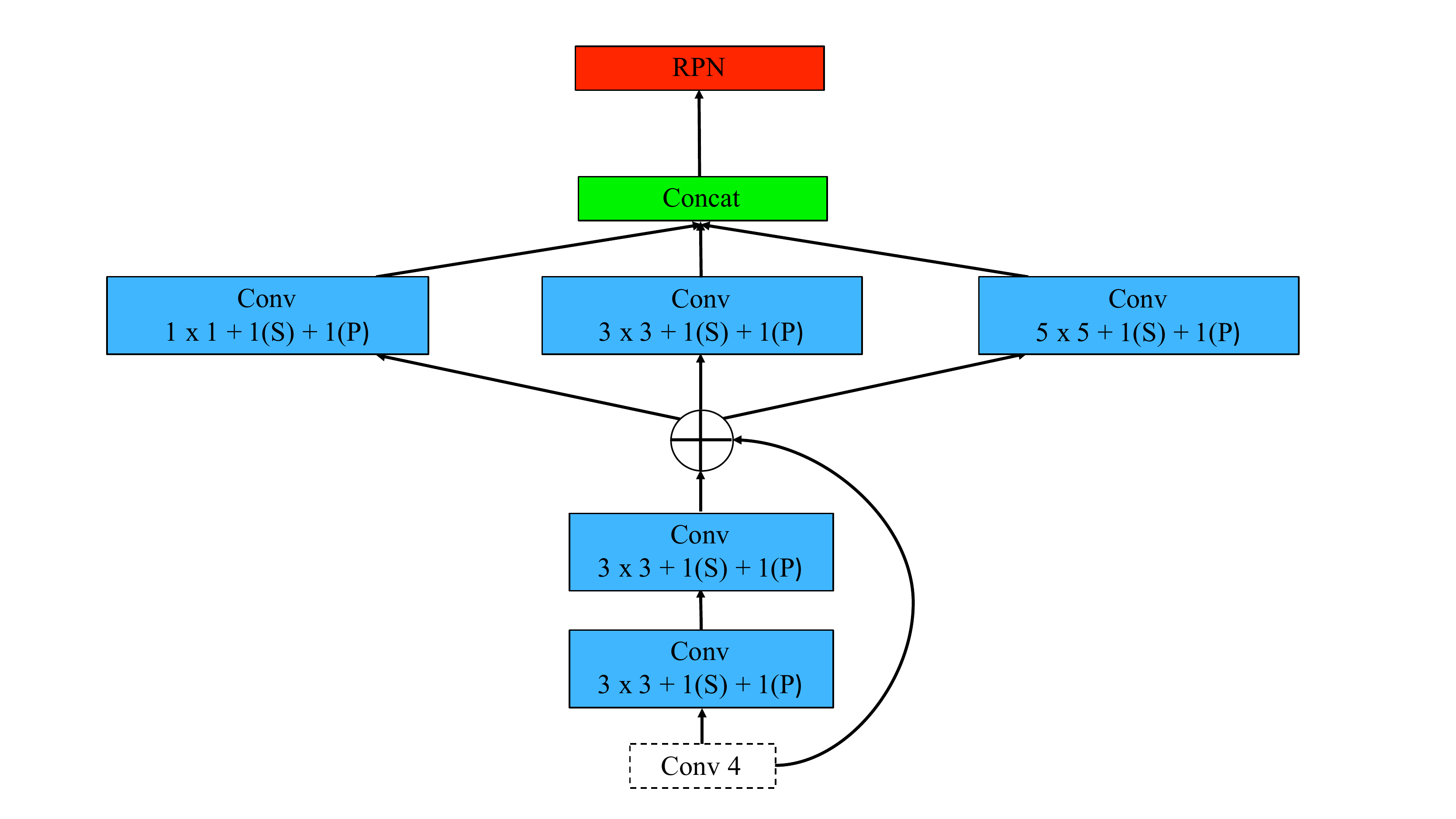}\\
  \caption{The proposed combination network architecture ZF$_{combin}$ combines ZF$_{ms}$ and ZF$_{res}$. The ZF$_{ml}$ model is not specifically combined based on our results that it doesn't improve detection precision much. }
  \label{Comb}
\end{figure}

\section{EXPERIMENTS}

\subsection{Experimental details}

\textbf{Network architecture}. The baseline network we use is Faster R-CNN with ZF net \cite{zeiler2014visualizing}, which is pre-trained by ImageNet. For the VOC dataset training, we trained 100000 iterations with a learning rate starting at 0.001 and dropping to 1/10 after 50000 steps. For KITTI dataset training, we trained 150000 iterations with a learning rate of 0.001 and a step size of 50000. We implement the CNNs based on the Caffe \cite{jia2014caffe} library. The baseline model is termed simply as ZF. The final network with a multi-layer representation, multi-scale representation, residual block embedding and tiny anchors are termed as ZF$_{ml}$, ZF$_{ms}$, ZF$_{res}$, and ZF$_{anchor}$. We finally combine these techniques to form a carefully designed network, which is termed as ZF$_{combin.}$.

\begin{table}
\centering
\begin{tabular}{|c|c|c|c|}
\hline
 & Multi-layer & Multi-scale & Residual block \\ \hline\hline
ZF & & &  \\ \hline
ZF$_{ml}$ & ${\surd}$  & &  \\ \hline
ZF$_{ms}$ &  & ${\surd}$ &   \\ \hline
ZF$_{res}$ &   &  &  ${\surd}$  \\ \hline
ZF$_{combin.}$ & ${\surd}$  & ${\surd}$  & ${\surd}$ \\  \hline
\end{tabular}
\caption{\label{tab:widgets1} Test Experiments on KITTI Dataset. We also add a ``combination'' model making use of all modifications in a single network.}
\label{tab:model}
\end{table}


\begin{table}
\centering
\begin{tabular}{|c|c|c|}
\hline
Model & Converged Epochs & AP Performance  \\ \hline\hline
ZF & 120,000 & 76.3 \\ \hline
ZF$_{ml}$ & 80,000 & 76.1 \\ \hline
ZF$_{ms}$ & 100,000 & 76.5 \\ \hline
ZF$_{res}$ & 100,000 & 76.9 \\ \hline
ZF$_{anchor}$ & 110,000 & 79.6 \\  \hline
\bf ZF$_{combin.}$ & \bf 120,000 & \bf 83.6 \\  \hline
\end{tabular}
\\
\caption{\label{tab:KITTI} Test Experiments on the KITTI Dataset. Here we see that the strongest single contribution is the anchor box adjustment with a strong contribution also provided by the residual network. The combination network is the strongest performer.}
\label{tab:FINAL}
\end{table}

\textbf{Training \& Testing Dataset}. The VOC2007 dataset \cite{everingham2007pascal} contains 2501 training images and 2510 testing images, including 6301 and 6307 objects, respectively. There are 21 classes in the dataset, such as aeroplane, bicycle, bird and so on. In KITTI dataset there are 7481 images in total and there are only 6684 images that contain cars. We randomly divide them into a training partition (5484 images) and the testing partition (1000 images) for six cross-validation folders. After training on each folder, we use the average AP as our AP for this model. In this study, only the car class in KITTI dataset is considered for simplicity.

\subsection{Baseline}
KITTI \cite{geiger2013vision} dataset is closer to a typical real-world scenario than the VOC data, as shown in Figures \ref{fig:sub1}, \ref{fig:sub2} and
Figure \ref{fig:sub3}. The KITTI dataset image size is 1392 $\times $ 512. Other datasets like VOC2007 and
ImageNet are carefully designed for general objection detection. Typically, each image contains one to two objects and
these objects occupy most of the image. On the other hand, the images in the KITTI dataset are taken from the viewpoint
of a vehicle on the road. In this case, cars appear anywhere in the image and perspective effects mean that the scale of
vehicles varies greatly.

\begin{table}
\centering
\begin{tabular}{ccc|c}
Method & Fine Tune & Test & MAP (\%) \\ \hline\hline
ZF  & VOC & VOC & 60.7 \\
ZF$_{ml}$ & VOC & VOC & 60.7 \\
ZF$_{ms}$ & VOC & VOC & 60.2 \\
ZF$_{res}$ & VOC & VOC & 61.2 \\
ZF$_{anchor}$ & VOC & VOC & 61.4 \\
\textbf{ZF$_{combin.}$} & \textbf{VOC} &  \textbf{VOC} & \textbf{61.6} \\
\\
\end{tabular}
\caption{\label{tab:widgets3} Test Experiments on the VOC Dataset. Here we see that the changes to the network design only provide minor improvements to performance due to lower scale-variance in the VOC dataset.}
\label{tab:VOC}
\end{table}



\par
We have three baseline results: the VOC dataset trained and tested Faster R-CNN detector, the VOC dataset trained and KITTI dataset tested Faster R-CNN detector, and the KITTI dataset trained and tested Fast R-CNN.
The VOC trained and tested network (See Table \ref{tab:VOC}) shows a little improvement via our proposed modifications. This is a consequence of the fact that our methods are designed for scale-variance and the VOC dataset has little scale variance. Furthermore, training via the VOC dataset reveals only a modest 0.50\% AP improvement for car detection.
However, the KITTI trained and tested Faster R-CNN detector shows a significant improvement to 76.3\%, as shown in Table \ref{tab:KITTI}. After fine tuned the Faster R-CNN network with KITTI dataset, we can see the accuracy increased by
26.3\% overall.

\subsection{Anchor Box Selection and Scale Performance}
\label{anchorboxresults}

\par
As shown in Table \ref{tab:VOC} and Table \ref{tab:KITTI}, the multi-scale AP on VOC2007 is 60.7\% which is also similar to the baseline result with only a 0.9\% improvement. For the KITTI dataset, the average performance among 6000 test images is improved from 76.3\% to 79.6\% using 5 scales. The limited improvement for the VOC dataset is again expected due to the low scale-variance. For the KITTI dataset, we can see that tiny objects can be more easily detected by multi-scale proposals than the original Faster R-CNN network. This is consistent with our expectation since tiny-anchor proposal are designed to give better region proposals for smaller vehicles.

\begin{figure}[ht]
   \centering
   \includegraphics[width=\linewidth]{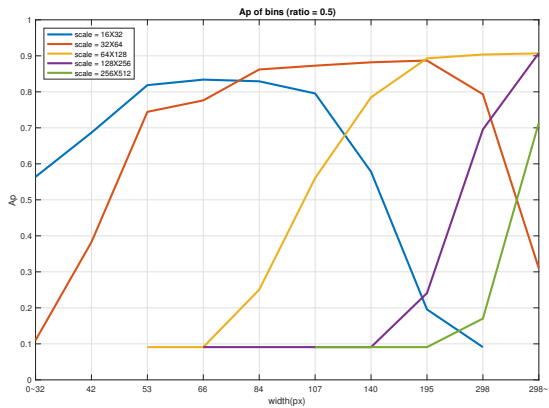}\\
   \caption{Anchor box scale vs detector performance for different sized objects. Detector performance is best around a limited scale range ``near'' to the anchor box scaling, performance drops quickly beyond a certain point after which a different anchor box provides the best chance of capturing an object. Interestingly, the AP curves are upward biased towards the larger scales. This can be attributed to the fact that detection performance improves quite markedly as the size of the object increases in the scene.} 
   \label{fig:anchorsmatter}
 \end{figure}

\par
\begin{figure}[t]
  \centering
  \includegraphics[width=\linewidth]{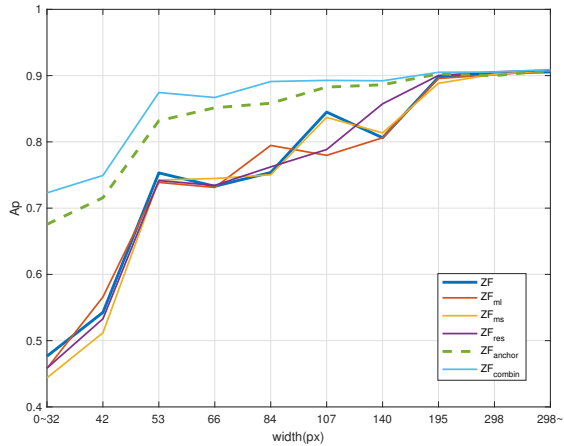}\\
  \caption{Detector performance vs object scales. For the larger object instances, all models perform at about the same level. However, our combination and anchor-box optimized models achieve a significant improvement for smaller object instances.}
  \label{fig:modelsatscale}
\end{figure}

Each scale's ability in detection car objects is also studied. The result is shown in Figure \ref{fig:anchorsmatter}. We can see the detection performance for a certain scale is the best around the object scale range near to the anchor box scale. In addition, we tested the performance of our various models against the images of specific scales to see where the AP performance gains were specifically realised as in Figure \ref{fig:modelsatscale}. This shows that the majority of the improvements comes from the handling of smaller scaled objects in the dataset.



\subsection{Model Choice and Scale Performance}





\begin{figure}[t]
  \centering
  \includegraphics[width=\linewidth]{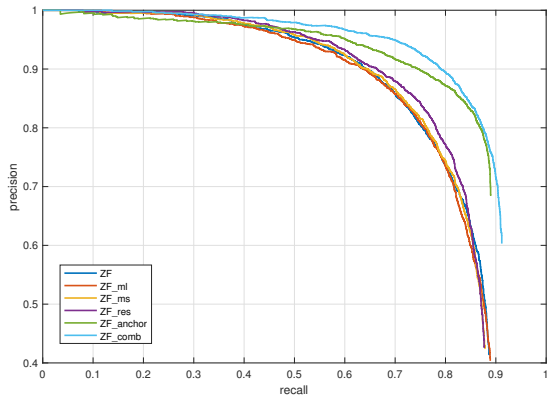}\\
  \caption{Precision-Recall Curve for Different Models. The combination model is the strongest due to its scale-variant design at anchor and network design level.}\label{prec_recall}
\end{figure}

\textbf{Multi-layer proposals} As shown in Table \ref{tab:VOC} and Table \ref{tab:KITTI}, the performance of the baseline and the multi-layer model on VOC2007 remains unchanged. For the Kitti dataset, the new model performance is improved from 64.0\% to 65.9\% after 10000 iterations but converges to a similar accuracy as the baseline ZF model. This shows that the concatenation of 4th and 5th layer of convolutions didn't improve the final accuracy of detection but decreases the required training time.
\par
\textbf{Multi-scale proposals} For the KITTI dataset, the multi-scale performance is significantly improved from 64.0\% to 66.3\% relative to the baseline model after 10000 iterations. However, the final performance converges to only a slightly higher AP. The final AP is 76.5\%. This is easy to understand since the multi-scale proposal has finer scaled features with the convolutional kernel combination of 1x1, 3x3 and 5x5. This provides better feature information across various scales than the baseline model.
\par
\textbf{Residual block embedding.} The detection performance of our residual block based method on the KITTI dataset is given in Table \ref{tab:VOC} and Table \ref{tab:KITTI}, respectively. This model improves the network performance by 0.6\% as the residual network embedding allows information from lower level convolutional layers (with finer/smaller scale features). The addition of two layers of convolutions and the residual connection makes the final feature map diverse, having high and mid-level features while also converging more quickly.

\textbf{The combination of the proposed techniques.} Since the proposed improvements can be incorporated into a single framework, we further perform an experiment based on a combination network as shown in Figure \ref{Comb}. As can be observed from Table \ref{tab:VOC}, the combined model works remarkably well on KITTI. It effectively validates our analysis on the KITTI dataset regarding small scale vehicles within the data. The final detection AP is significantly improved to 83.6\%.


\section{CONCLUSION}
In this study, we explored the application of the Faster R-CNN framework in a vehicle detection task. As Faster R-CNN is particularly designed for general object detection for objects of a particular scale-distribution and frequency, it requires some adaptation in order to work well on a vehicle detection task. This is primarily attributed to the need for a vehicle detection method to match the significant variations in the natural distribution of target object scale, position, and frequency. To address this issue, we proposed several modifications on the network architecture's convolutional layers and region proposal selections. We combined multiple level features to obtain better performance across scales while adding additional anchor box suggestions. Our experiments on KITTI dataset improve detection performance relative to our baseline by 7.3\%. We anticipate a strong future for full-image convolution methods based on Faster-RCNN for real-world problems such as the KITTI Vehicle Detection. Combined with CNN network designs able to capture features at a large variety of scales, such as shown in this paper, we believe these methods will displace the traditional exhaustive scale-space search approach for real-world real-time object detection.

\addtolength{\textheight}{-13cm}   









\bibliographystyle{IEEEtran} 
\bibliography{bib_data}

\begin{thebibliography}{10}
\providecommand{\url}[1]{#1}
\csname url@rmstyle\endcsname
\providecommand{\newblock}{\relax}
\providecommand{\bibinfo}[2]{#2}
\providecommand\BIBentrySTDinterwordspacing{\spaceskip=0pt\relax}
\providecommand\BIBentryALTinterwordstretchfactor{4}
\providecommand\BIBentryALTinterwordspacing{\spaceskip=\fontdimen2\font plus
\BIBentryALTinterwordstretchfactor\fontdimen3\font minus
  \fontdimen4\font\relax}
\providecommand\BIBforeignlanguage[2]{{%
\expandafter\ifx\csname l@#1\endcsname\relax
\typeout{** WARNING: IEEEtran.bst: No hyphenation pattern has been}%
\typeout{** loaded for the language `#1'. Using the pattern for}%
\typeout{** the default language instead.}%
\else
\language=\csname l@#1\endcsname
\fi
#2}}

\bibitem{lin2014microsoft}
T.-Y. Lin, M.~Maire, S.~Belongie, J.~Hays, P.~Perona, D.~Ramanan,
  P.~Doll{\'a}r, and C.~L. Zitnick, ``Microsoft coco: Common objects in
  context,'' in \emph{European Conference on Computer Vision}.\hskip 1em plus
  0.5em minus 0.4em\relax Springer, 2014, pp. 740--755.

\bibitem{everingham2010pascal}
M.~Everingham, L.~Van~Gool, C.~K. Williams, J.~Winn, and A.~Zisserman, ``The
  pascal visual object classes (voc) challenge,'' vol.~88, no.~2, pp. 303--338,
  2010.

\bibitem{russakovsky2015imagenet}
O.~Russakovsky, J.~Deng, H.~Su, J.~Krause, S.~Satheesh, S.~Ma, Z.~Huang,
  A.~Karpathy, A.~Khosla, M.~Bernstein, \emph{et~al.}, ``Imagenet large scale
  visual recognition challenge,'' \emph{International Journal of Computer
  Vision}, vol. 115, no.~3, pp. 211--252, 2015.

\bibitem{girshick2014rich}
R.~Girshick, J.~Donahue, T.~Darrell, and J.~Malik, ``Rich feature hierarchies
  for accurate object detection and semantic segmentation,'' in
  \emph{Proceedings of the IEEE conference on Computer Vision and Pattern
  Recognition}, 2014, pp. 580--587.

\bibitem{girshick2015fast}
R.~Girshick, ``Fast r-cnn,'' in \emph{Proceedings of the IEEE International
  Conference on Computer Vision}, 2015, pp. 1440--1448.

\bibitem{fasterrcnn}
S.~Ren, K.~He, R.~Girshick, and J.~Sun, ``Faster r-cnn: Towards real-time
  object detection with region proposal networks,'' in \emph{ANIPS}, 2015, pp.
  91--99.

\bibitem{redmon2016you}
J.~Redmon, S.~Divvala, R.~Girshick, and A.~Farhadi, ``You only look once:
  Unified, real-time object detection,'' in \emph{Proceedings of the IEEE
  Conference on Computer Vision and Pattern Recognition}, 2016, pp. 779--788.

\bibitem{geiger2013vision}
A.~Geiger, P.~Lenz, C.~Stiller, and R.~Urtasun, ``Vision meets robotics: The
  kitti dataset,'' \emph{The International Journal of Robotics Research},
  vol.~32, no.~11, pp. 1231--1237, 2013.

\bibitem{viola2001rapid}
P.~Viola and M.~Jones, ``Rapid object detection using a boosted cascade of
  simple features,'' in \emph{Proceedings of the IEEE Conference on Computer
  Vision and Pattern Recognition}, vol.~1, 2001, pp. I--I.

\bibitem{szeliski2010computer}
R.~Szeliski, \emph{Computer vision: algorithms and applications}.\hskip 1em
  plus 0.5em minus 0.4em\relax Springer Science \& Business Media, 2010.

\bibitem{dalal2005histograms}
N.~Dalal and B.~Triggs, ``Histograms of oriented gradients for human
  detection,'' in \emph{Proceedings of the IEEE Conference on Computer Vision
  and Pattern Recognition}, vol.~1, 2005, pp. 886--893.

\bibitem{li2015convolutional}
H.~Li, Z.~Lin, X.~Shen, J.~Brandt, and G.~Hua, ``A convolutional neural network
  cascade for face detection,'' in \emph{CVPR}, 2015, pp. 5325--5334.

\bibitem{verma2015pedestrian}
A.~Verma, R.~Hebbalaguppe, L.~Vig, S.~Kumar, and E.~Hassan, ``Pedestrian
  detection via mixture of cnn experts and thresholded aggregated channel
  features,'' in \emph{Proceedings of the IEEE International Conference on
  Computer Vision Workshops}, 2015, pp. 163--171.

\bibitem{krizhevsky2012imagenet}
A.~Krizhevsky, I.~Sutskever, and G.~E. Hinton, ``Imagenet classification with
  deep convolutional neural networks,'' in \emph{Advances in neural information
  processing systems}, 2012, pp. 1097--1105.

\bibitem{uijlings2013selective}
J.~R. Uijlings, K.~E. Van De~Sande, T.~Gevers, and A.~W. Smeulders, ``Selective
  search for object recognition,'' \emph{International journal of computer
  vision}, vol. 104, no.~2, pp. 154--171, 2013.

\bibitem{zitnick2014edge}
C.~L. Zitnick and P.~Doll{\'a}r, ``Edge boxes: Locating object proposals from
  edges,'' in \emph{European Conference on Computer Vision}.\hskip 1em plus
  0.5em minus 0.4em\relax Springer, 2014, pp. 391--405.

\bibitem{zeiler2014visualizing}
M.~D. Zeiler and R.~Fergus, ``Visualizing and understanding convolutional
  networks,'' in \emph{European conference on computer vision}.\hskip 1em plus
  0.5em minus 0.4em\relax Springer, 2014, pp. 818--833.

\bibitem{zhang2016faster}
L.~Zhang, L.~Lin, X.~Liang, and K.~He, ``Is faster r-cnn doing well for
  pedestrian detection?'' in \emph{European Conference on Computer
  Vision}.\hskip 1em plus 0.5em minus 0.4em\relax Springer, 2016, pp. 443--457.

\bibitem{fan2016closer}
Q.~Fan, L.~Brown, and J.~Smith, ``A closer look at faster r-cnn for vehicle
  detection,'' in \emph{Intelligent Vehicles Symposium (IV), 2016 IEEE}.\hskip
  1em plus 0.5em minus 0.4em\relax IEEE, 2016, pp. 124--129.

\bibitem{sun2014deep}
Y.~Sun, X.~Wang, and X.~Tang, ``Deep learning face representation from
  predicting 10,000 classes,'' in \emph{Proceedings of the IEEE Conference on
  Computer Vision and Pattern Recognition}, 2014, pp. 1891--1898.

\bibitem{szegedy2015going}
C.~Szegedy, W.~Liu, Y.~Jia, P.~Sermanet, S.~Reed, D.~Anguelov, D.~Erhan,
  V.~Vanhoucke, and A.~Rabinovich, ``Going deeper with convolutions,'' in
  \emph{Proceedings of the IEEE Conference on Computer Vision and Pattern
  Recognition}, 2015, pp. 1--9.

\bibitem{he2016deep}
K.~He, X.~Zhang, S.~Ren, and J.~Sun, ``Deep residual learning for image
  recognition,'' in \emph{Proceedings of the IEEE Conference on Computer Vision
  and Pattern Recognition}, 2016, pp. 770--778.

\bibitem{jia2014caffe}
Y.~Jia, E.~Shelhamer, J.~Donahue, S.~Karayev, J.~Long, R.~Girshick,
  S.~Guadarrama, and T.~Darrell, ``Caffe: Convolutional architecture for fast
  feature embedding,'' in \emph{Proceedings of the 22nd ACM international
  conference on Multimedia}.\hskip 1em plus 0.5em minus 0.4em\relax ACM, 2014,
  pp. 675--678.

\bibitem{everingham2007pascal}
M.~Everingham, A.~Zisserman, C.~K. Williams, L.~Van~Gool, M.~Allan, C.~M.
  Bishop, O.~Chapelle, N.~Dalal, T.~Deselaers, G.~Dork{\'o}, \emph{et~al.},
  ``The pascal visual object classes challenge 2007 (voc2007) results,'' 2007.

\end{thebibliography}

\end{document}